# Know Your Personalization: Learning Topic level Personalization in Online Services

Anirban Majumder
Bell Labs Research, Bangalore, India
anirban.majumder@alcatel-lucent.com

Nisheeth Shrivastava
Bell Labs Research, Bangalore, India
nisheeth.shrivastava@alcatel-lucent.com

## ABSTRACT

Online service platforms (OSPs), such as search engines, news-websites, ad-providers, etc., serve highly personalized content to the user, based on the *profile* extracted from his history with the OSP. Although personalization (generally) leads to a better user experience, it also raises privacy concerns for the user—he does not know what is present in his profile and more importantly, what is being used to personalize content for him. In this paper, we capture OSP's personalization for an user in a new data structure called the *personalization vector* ($\eta$), which is a weighted vector over a set of *topics*, and present techniques to compute it for users of an OSP.

Our approach treats OSPs as black-boxes, and extracts $\eta$ by mining only their *output*, specifically, the *personalized* (for an user) and *vanilla* (without any user information) contents served, and the differences in these content. We believe that such treatment of OSPs is a unique aspect of our work, not just enabling access to (so far hidden) profiles in OSPs, but also providing a novel and practical approach for retrieving information from OSPs by mining differences in their outputs.

We formulate a new model called Latent Topic Personalization (LTP) that captures the personalization vector into a learning framework and present efficient inference algorithms for it. We do extensive experiments for search result personalization using both data from real Google users and synthetic datasets. Our results show high accuracy (R-pre = 84%) of LTP in finding personalized topics. For Google data, our qualitative results show how LTP can also identifies *evidences*—queries for results on a topic with high $\eta$ value were re-ranked. Finally, we show how our approach can be used to build a new Privacy evaluation framework focused at end-user privacy on commercial OSPs.

## 1. INTRODUCTION

Personalization is being used by most online service platforms (OSPs) such as search, advertising, shopping, etc. The goal is to lure users by offering a better service experience customized to their individual interests. A popular trend is to employ profile based personalization, where OSPs build extensive profile for the user (based his past interactions search queries, browsing history, links shared, etc.) and personalize the content based on this profile. Several popular services employ such personalization, e.g. search[1], movie recommendations[2], etc.

While OSPs definitely track rich user histories, they can infer a great deal more by mining this raw data. Informally speaking, OSPs can determine users interests and biases on different categories, which can then be used (along with his history) for personalization. For example (see [19] for details), Google is shown to have inferred users political affiliations (republican or democratic), and use it to re-ranked results.

For an user, this raises a significant privacy concern—he does not know what was tracked in his history, what has been inferred, and more importantly, is currently being used to personalize content. Moreover, as both the personalization techniques and the data they operate on are the key differentiators of these OSPs (their *secret sauce*), they do not reveal either of them, making it even harder for an user to understand how personalization is done for him.

In this paper, we aim at extracting an user's profile from the OSP. We model an user's profile as a weighted personalization vector over topics, where the weight on a topic indicates his interest in it (higher means more interested). Informally, a topic is any concept or phenomenon that the user could be interested in, e.g. a specific sport, a preference over cuisine, favorite author, movie genre, etc.[3]

Our goal is not to reverse-engineer the OSPs inference algorithms. In fact, we treat the OSP as a black-box. We assume that we only have access to their output, which is basically the (personalized) content served by them to the user on different urls[4]. The key idea is to get both *personalized content* (served for the user) and *vanilla content* (served for a new/not logged in user) for the same url from the OSP, and determine the topics of personalization based these two content and the differences among them.

---

[1]http://privacy.microsoft.com/en-us/Bing.mspx; http://support.google.com/websearch/bin/answer.py?answer=1710607.
[2]Netflix: http://www.netflixprize.com
[3]More specifically, we define it as a distribution over bag of words as common in the topic modeling literature (see Section 3 for details).
[4]The url could point to a static page, e.g. reviews and other information on a movie, or dynamically generated, e.g. search results.

The profiles in OSPs have remained opaque so far, with little knowledge of user profiles hidden in them. Our paper provides a novel approach to crack this problem, giving insights into the user profiles without the knowledge of the inference techniques or the history of the user. We believe that this aspect of comparing the differences in output to extract the hidden personalized topics is unique to our paper and opens a new directions in privacy research that can be aimed at commercial OSPs.

Let us consider the case of a search engine. For any query, we can get the personalized and vanilla results by making the query from a browser with and without logging in, respectively. These results are basically two ranked lists with some urls in the latter moved up or down in the former, based on the user profile. We study these movements over multiple queries and determine the most likely topics of interest for the user that can best explain them.

For the remaining of this paper we will talk only about search result personalization. However, our techniques can be easily extended to any service where a) we can observe both vanilla and personalized content and b) we can get a ranked ordering of the content. For example, we can apply it to movie recommendation (in say Netflix) based on the personalized (and vanilla) ranked list of related movies presented when on a web-page of a particular movie.

## 1.1 Search Personalization and Re-ranking

Although the exact details of personalization for many popular services of today's web remain a mystery, recent works in the web-search community have thrown some light into the intricacies of search engine personalization [8, 22, 17]. These techniques vary considerably in terms of their description and complexity but the common underlying theme for them is to first populate the vanilla result using the semantics of the query string and then personalize it by re-arranging the items in this list, using the profile information. Therefore, conceptually, the vanilla and personalized responses are re-ordering of the same set of items. We take advantage of this re-ranking of results to determine the topics present in the user's profile with the OSP.

The restriction of re-ranking over the *same* urls is useful for exposition of our solution approach, but can be easily lifted by simply adding the extra urls in one list to the end of the other list[5]. The important point is that personalization, by definition, will affect ranks of results shown, which is what we use in this paper.

Note that these topics may not be explicitly maintained at the OSP; in fact they could be using something completely unrelated to our definition of topics to model the user profile. Our paper hinges on the intuition that an user's interests with most OSPs can be captured by a set of topics that he is interested in. And any OSP that personalizes results based on his interests must give higher preference to results matching these topics. Thus our approach of finding topic-level personalization is fairly generic—working on OSPs who do not necessarily have topic-based profiles of users and without the knowledge of the profiling algorithms they use.

A alternate competitive approach to recreate the user profile could be by mining the *input* to the OSP (i.e. user's history)[8, 22, 21]. However, this approach has several shortcomings compared to us. One, it is very hard to catch up to the commercial techniques used by OSPs that are usually more advanced and rapidly evolving. Two, due to proprietary nature of OSPs, it is not clear what algorithm or even what part of the history is being used by them (e.g. most personalization is done using only recent history, but it is not clear as to how recent for each user). In other words, with any profiling tool, there is no certainty that it can infer all that the OSP has. Finally, in many cases the history information may not be available publicly (i.e. while a Google user's search history is available, past ads served are not), limiting the effectiveness of these approaches. In contrast, our approach is agnostic to OSP's personalization scheme and can work even when the history is not public.

## 1.2 A new privacy preserving framework

The topics of personalization for an user can be utilized in building a novel privacy evaluation and prevention toolkit, that we describe next. The toolkit presents user with the topics his profile is personalized on and ask him to determine, based on his personal judgment, whether some of these topics are *sensitive*[6]. Now, a topic which is both sensitive and has high personalization score can be deemed a privacy leak, as the OSP is using an user's data in a way he does not agree with. These leaks for a user can now be detected and monitored over time, and in several cases can also be plugged, e.g. by undoing the re-ranking on sensitive topics or simply served the vanilla content. These ideas are currently being developed into a privacy preserving toolkit in which the techniques developed in this paper are a key component.

## 1.3 Our Contributions

The main contributions of the paper are as follows.

- We propose a new direction in privacy research that enables users in getting a glimpse of their profile information being used by commercial OSPs to serve personalized content. We formally capture this information as a topic-level personalization vector that provides a concise and accurate summary of the user profile.

- We propose a novel way to compute this topic-level personalization based on the personalized and vanilla content served by OSPs. This formulation treats the services as a black box and hence can work with a variety of online services. We believe that this is a unique aspect of our work and can open a new direction for privacy research by enabling access to (so far hidden) profile information in OSPs.

- We present a probabilistic model (named Latent Topic Personalization, or LTP) that captures the intuition behind our approach. LTP is both expressive and leads to computationally efficient inference algorithms (LTP-INF and LTP-EM) that find the personalization vector on real datasets.

---

[5]In our experiments with Google, only 15% of personalized results contain any extra result compared to vanilla, and even these contain on average only 14% extra urls (or, 1.4 urls for an avg. result size of 10).

[6]The definition of sensitive, informally stated as any topic he find uncomfortable getting personalization on, can vary across users and could include health conditions, financial, sexual preferences, etc.

- Our experiments with synthetic dataset using state-of-the-art personalization engine show that LTP can learn the personalization parameters very accurately, getting on average 84% precision in learning personalized topics.

- We perform experiments on a novel real-life dataset containing the personalized and vanilla query results collected from 10 Google users. We also demonstrate how our techniques can be used to find the *evidence* of personalization which can be very helpful in user facing tools (see Section 1.2).

## 2. RELATED WORK

**Search personalization:** A large body of work exists on personalizing search results using user-profiles [8, 22, 17], that collectively give overwhelming evidence of its benefits. More recently, researchers have also explored creating profiles using topic models [21] and other textual information [23]. These works are not competitors of our paper, but rather serve as a motivation for us, as they highlight existence and importance of profiles in the state-of-the-art in personalization.

Another body of work explores short-term and session based personalization [1, 8], that personalize based on user's current *intention*, based on his recent history or session. While such approach is not aligned with our idea, there are two important points to note—a) they do no imply profile-based personalization does not happen, rather, they are typically used in conjunction with each other [1, 13], and b) since they are applicable only during a session, it is easy to remove their affect by making sure no coherent session is tracked during our data collection (by doing queries randomly and multiple times while matching results).

Researchers have also found that personalization is not always beneficial and have proposed various approaches, such as click-entropy [24, 26], dynamic user interests [13] and query difficulty [30], to filter queries that should not be personalized (irrespective of user's profile). Such filtering is very hard replicate in our approach since the output may not contain any information to model them. We therefore allow for existence of this hidden process in our model via a latent variable deciding (randomly) if personalization happens on a query (see Section 4.1 for details).

**Topics Models:** Although topic models are clearly a popular tool for processing textual information and have been also used in personalization, there is no work to our knowledge that models the differences in two documents (or two ranked set of documents) as us. A recent work by Bischof et. al.[2] comes close—they find exclusive topics (that are sufficiently different from each other) so that the documents can be classified into non-overlapping hierarchy. While this also involves finding topics which are present in some documents and not in others, it is still very different from our approach of finding a consistent (may not be exclusive) set of personalized topics that can differentiate personalized and vanilla content.

**Privacy:** Finally, our problem stems from the general area of user privacy. Various studies have highlighted problems of privacy in information leaks from OSPs[11, 16, 10]. Korolova et. al.[10] showed how targeted ads can pin-point individual users in Facebook, Mao et. al.[16] analyzed tweets to find vacation plans, medical conditions etc. for real users. However, these studies are focused on finding instances of privacy leaks from the entire OSP network and do not help users understand leaks in their own account. Other approaches of privacy preserving personalization aim at building a system from the scratch that ensures certain norms are preserved in the personalized output, e.g. grouping user profiles [28, 29] to preserve k-anonymity or making a differentially private recommender system[14]. Recently, Chen et. al.[6] presented a more user centric approach that gives user control over fine grained categories (represented as a fixed hierarchical taxonomy) which they want personalization on. These techniques however require users switch to these new systems from their existing OSPs, which is not practical, while we aim at finding personalization in existing OSPs.

## 3. PROBLEM FORMULATION

In this section, we introduce our notations and define the technical problem that we consider in this paper.

### 3.1 Notation

Let $I = \{i_1, i_2, \cdots\}$ be the universe of all the *items* being present at the personalization server, where, an item might represent a url (for search engines like Google, Bing etc.), a product web-page (for e-commerce sites like Amazon, Net-Flix etc.) or an advertisement (for ad servers). For a query $q$, let $\pi_q$ and $\sigma_q$ denote the personalized and vanilla lists of content. In the following discussion, we will often drop the subscript $q$, when the query is understood from the context.

As mentioned earlier, both $\pi$ and $\sigma$ are treated as permutations over a set of items[7] $I' \subset I$. Technically, a ranking/permutation[8] is a bijection from a set to itself. For any permutation $\pi$, $\pi(i)$ denotes the item assigned to rank $i$, hence $\pi = (\pi(1), \pi(2), \cdots)$. $\pi^{-1}(d)$ denotes the rank $i$ of an item $d \in I$ in $\pi$ such that $\pi(i) = d$. For any two permutations $\pi$ and $\sigma$, we use the notation $\sigma^{-1}(\pi(i))$ to denote the rank of the item $\pi(i)$ in $\sigma$. Observe that $\pi^{-1}(\pi(i)) = i$. We use $S_n$ to denote the set of all permutation of $n$ items.

We assume that there are $T$ topics $\{\beta_1, \beta_2, \cdots, \beta_T\}$ in our system where each topic $\beta_k$ is defined as a multinomial distribution over a fixed vocabulary $V$. For each word $w \in V$, we have a parameter $\beta_{k,w} = \Pr(w \mid \beta_k)$ such that $\sum_{w \in V} \beta_{k,w} = 1$. Each item[9] $i \in I$ is represented by its topic-map $\theta_i$ which is a multinomial distribution over the set of topics. By inspecting each component of $\theta_i$, one can infer how related the item is to a particular topic.

We now describe our representation of topic-level user profile information. For each user $u$ and topic $\beta_k \in \beta$, we associate a variable $\eta_{u,k} \in R$. It captures the importance of $\beta_k$ (more relevant topics have higher values) for serving personalized content to $u$. The complete profile information (we name it as *latent personalization vector*) is denoted by $\eta_u = (\eta_{u,1}, \eta_{u,2}, \cdots, \eta_{u,T})$. We often drop the subscript $u$ and refer to it simply as $\eta$ whenever the user is understood from the context.

---

[7]Strictly speaking, they may not contain exactly same set of items, but it is normally the case. E.g. in our experiments with Google, personalized results are identical to vanilla for 85% of queries and contain only avg. 14% extra items on the remaining. These extra items can be handled easily by adding them to the end of personalized (or vanilla) list.

[8]We often use them interchangeably.

[9]Specifically, the textual content or meta-data of the item.

## 3.2 Problem

Our strategy to learn the personalization vector $\eta$ is to repeatedly frame queries to the server and observe the difference between its vanilla and personalized responses. For a given user $u$, we first sign-in to her account and submit a query to the server. This gives the server an opportunity to personalize the result by using $u$'s profile information and through this process, we obtain the personalized response $\pi$. Next, we submit the same query in an anonymized form, by removing all cookies from the http request, thus removing all account details (but keeping all other parameters same such as IP address, User-Agent, etc.). This time the server sends back the vanilla response $\sigma$. We expect that as this process is repeated many times, the cumulative difference between these two responses will become statistically significant and contain substantial evidence of $\eta$. In this paper, we study the following problem: *Given pairs of query results $(\sigma_1, \pi_1), (\sigma_2, \pi_2) \cdots (\sigma_m, \pi_m)$, how do we learn the latent personalization vector $\eta$, for a given user?*

**Non-profile factors** Although personalization normally yields its benefits by presenting more relevant results to the users, it is also known to be less effective and even detrimental in many cases. For example, while personalizing results are known to work well for short and ambiguous queries [25] where user searching same query may be looking for completely different things, for common and specific queries two users with very different profiles are normally looking for the same information and are satisfied with the same (ordering of) results. In such cases, even though user's profile implies re-ranking, the server may decide not to personalize. This creates a problem for our approach as a search engine's decision whether to personalize the result of a search query or not, is influenced not only by the topical content of the query result, but also through other filtering processes that are hidden from us.

We take care of this in our model by introducing a latent parameter that, during training phase, filters out such inexplicable events and reduces the noise in the personalization vector. In our experiments with the Google dataset, we found several instances of queries with results at higher ranks having higher "scores" (see Section 4 for definition of scores) the ones at lower ranks, that were not personalized, while another query with similar scores was personalized. Without this latent parameter, these instances would have reduced the effectiveness of learning $\eta$.

## 4. LTP MODEL

The goal of topic-based personalization learning is to capture the following information: topics on which personalization takes place and a weight vector corresponding to the degree of personalization on these topics. In addition, the approach has to scale with large number of queries. To meet these objectives, we first propose Latent Topical Personalization model (LTP) to study the problem from a bayesian perspective. Following that, we develop efficient variational inference and estimation techniques for learning the parameters of this model.

### 4.1 Model Description

We now formally describe the proposed LTP model. LTP models (Figure 1) both topics and personalization. It involves a *topic block* to model the topical content creation of the items and a *personalization block* to model the personalized responses (i.e. $\pi_1, \pi_2, \cdots, \pi_m$).

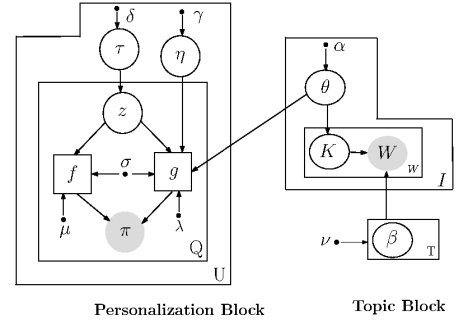

**Figure 1: Graphical model representation of LTP.**

**Topic Block** The topic block follows the description of standard topic models (c.f. LDA [3]) and we present it here for the sake of completeness. The generative process for the topic block is as follows

- For each topic $\beta_k, k = 1, 2 \cdots, T$
  1. Sample $\beta_k \sim Dirichlet(\nu)$.
- For each item $i \in I$
  1. Sample its topic-map $\theta_i \sim Gaussian(0, diag(\alpha^2))$.
  2. For each word position $j = 1 \cdots n_i$ for item $i$
     (a) Sample topic $K_{i,j}$ with $\Pr(K_{i,j} = k) \propto e^{\theta_{i,k}}$.
     (b) Sample word $W_{i,j} \sim Multinomial(\beta_{K_{i,j}})$.

The joint distribution for the topic-block can be written as

$$p(\theta, K, W, \beta \mid \alpha, \nu) = \prod_{i \in I} p(\theta_i \mid \alpha) \cdot \prod_{k=1}^{T} p(\beta_k \mid \nu)$$
$$\prod_{i \in I} \prod_{j=1}^{n_i} p(K_{i,j} \mid \theta_i) \cdot p(W_{i,j} \mid K_{i,j}, \beta_{1 \ldots T}) \quad (1)$$

**Personalization Block** Our design of the personalization block is little more involved. The main difficulty stems from the non-profile based factors, which may lead to no re-ranking of results even when the user profile (i.e. $\eta$) indicates personalization should happen. In LTP, we achieve it by introducing a latent switch variable $z$ (refer to Figure 1). Independently, for each query, we sample $z$, governed by a prior parameter $\tau$ and based on its value decide whether to allow topical personalization or not. The parameter $\tau$ is user-specific and controls the rate at which topical personalization takes place (for that user).

Based on the value of $z$, we pick a probability distribution over permutations and sample $\pi$ from it. Probabilistic models on permutations have recently been applied to solve various problems related to ranking [20]. Probability distributions defined over permutations can be broadly categorized into two types—*distance based* and *score based*. In a distance based model [15], the probability of a permutation is defined according to its distance from a central permutation. They have rich expressive power as they can incorporate a wide variety of distance functions over permutations but are, in general, computationally inefficient.

| σ | 2 | 3 | 1 |
|---|---|---|---|
| π | 3 | 1 | 2 |

| Items | 1 | 2 | 3 | π | | |
|---|---|---|---|---|---|---|
| 1st Stage | $e^{-2}$ | $e^0$ | $e^{-1}$ | 3 | | |
| 2nd Stage | $e^{-1}$ | $e^1$ | | 3 | 1 | |
| 3rd Stage | | 1 | | 3 | 1 | 2 |

$$f(\pi \mid \sigma) = 0.03$$

**Figure 2: An example illustrating the steps of $f$. We have assumed $\mu = 1$. At each stage, the actual outcome is marked in blue and the most likely outcome is marked in red.**

Score based models [12], on the other hand, are very efficient as they divide permutation construction into stages and assign scores on each stage such that the final probability is a combination (multiplication) of stage-wise scores. However, being defined as a specific function over scores, they have limited expressive power e.g. they can not take into account any central permutation in the generative process. For LTP, we have a central permutation (vanilla list $\sigma$) and want to model $\pi$ as being generated from it. Further, as explained later, we define scores on items as a function $\eta$. Therefore, we need a model which combines the notion of distance with scores and is computationally efficient.

The probability distribution $f$ (Figure 1) is a process for generating the personalized response $\pi$, and is decomposed into sequential stages. Observed that (see Figure 1) this process is activated only if $z = 0$, thereby, implying no topical personalization should happen. In the first stage, we pick the item $\pi(1)$ with probability $\frac{\exp(\mu(1-\sigma^{-1}\pi(1)))}{\sum_{j \geq 1}\exp(\mu(1-\sigma^{-1}\pi(j)))}$. Note that this probability is maximum when the two permutations agree with the first position i.e. $\pi(1) = \sigma(1)$. However, if we happen to pick some other item i.e. $\pi(1) \neq \sigma(1)$, then for the second stage, the most likely outcome is to bring back the item $\sigma(1)$ and put it at the second position of $\pi$ i.e. $\pi(2) = \sigma(1)$.

In general, in the $k^{th}$ stage, the probability of selecting $\pi(k)$ is $\frac{\exp(\mu(k-\sigma^{-1}\pi(k)))}{\sum_{j \geq k}\exp(\mu(k-\sigma^{-1}\pi(k)))}$. Intuitively, at each stage $k$, the model determines the items among $\sigma(1), \sigma(2), \cdots, \sigma(k-1)$ which are not yet sampled by $f$ and assigns higher probability on picking them. In Figure 2 gives an example of this sampling process.

Considering all the stages, we obtain the overall probability of sampling $\pi$ which is given by the following expression

$$f(\pi \mid \sigma, \mu) = \prod_i \left( \frac{\exp(\mu(i-\sigma^{-1}\pi(i)))}{\sum_{j \geq i}\exp(\mu(i-\sigma^{-1}\pi(j)))} \right) \quad (2)$$

It can be shown that $f$ is a valid probability distribution i.e. $f(\pi \mid \sigma, \mu) \geq 0$ for all $\pi \in S_n$ and $\sum_\pi f(\pi \mid \sigma, \mu) = 1$. The parameter $\mu$ controls the spread of the distribution i.e. if $\mu \to 0$ then $f$ converges to the uniform distribution over $S_n$; otherwise, for $\mu > 0$ the distribution is concentrated around $\sigma$. We assume $\mu \geq 1$.

We now describe our next permutation model $g$ that captures the topic-level personalization which is invoked only if $z = 1$. Model $g$ is also decomposed into sequential stages and at each stage uses both the central permutation $\sigma$ and a set of scores, to determine $\pi$. Each item $d \in I$ is assigned a score $\eta^T \theta_d$. In the $i^{th}$ stage, $g$ selects the item $\pi(i)$ with probability

$$\frac{\exp(\lambda\eta^T\theta_{\pi(i)} + (1-\lambda)(i-\sigma^{-1}\pi(i)))}{\sum_{j \geq i}\exp(\lambda\eta^T\theta_{\pi(j)} + (1-\lambda)(i-\sigma^{-1}\pi(j)))}$$

The working principle for $g$ is similar to $f$, except that it now allows for deviations from $\sigma$ only if it is explained by the scores. Parameter $\lambda$ is tuned to adjust the relative importance of the scores and the central permutation $\sigma$. For example, if $\lambda = 0$ then the scores are ignored and if $\lambda = 1$ then the central permutation does not play any role. We treat $0 \leq \lambda \leq 1$ as a free parameter whose value needs to be learned from the data. The overall probability of sampling $\pi$ is given by

$$g(\pi \mid \eta; \sigma, \lambda, \theta) = \prod_i \left( \frac{\exp(\lambda\eta^T\theta_{\pi(i)} + (1-\lambda)(i-\sigma^{-1}\pi(i)))}{\sum_{j \geq i}\exp(\lambda\eta^T\theta_{\pi(j)} + (1-\lambda)(i-\sigma^{-1}\pi(j)))} \right)$$

It can be verified that $g$ is also a valid probability distribution.

The generative process for the personalization block can be described as

- For each user $u$
  1. Sample $\tau \sim Beta(\delta, \delta)$.
  2. Sample $\eta \sim Gaussian(0, diag(\gamma^2))$
- For each query $q_i, i = 1, 2, \cdots, m$
  1. Sample $z_i \sim Bernoulli(\tau)$ to decide whether to allow topical personalization.
  2. If $z_i = 1$, sample $\pi_i \sim g(\cdot \mid \sigma_i, \lambda, \theta, \eta)$.
  3. Else, sample $\pi_i \sim f(\cdot \mid \sigma_i, \mu)$.

The joint distribution for the personalization block can be written as

$$p(\pi, z, \tau, \eta \mid \theta; \gamma, \delta, \mu, \lambda, \sigma) = p(\eta \mid \gamma) \cdot p(\tau \mid \delta)$$
$$\prod_{i=1}^{m} p(z_i \mid \tau) \cdot g(\pi_i \mid \sigma_i, \lambda, \theta, \eta)^{z_i} f(\pi_i \mid \sigma_i, \mu)^{1-z_i} \quad (3)$$

Finally, the full joint distribution for LTP can be obtained by multiplying Equations 1 and 3. We treat the parameters $\nu, \alpha, \delta, \gamma$ as constant and do not consider learning them. However, the parameters $\mu$ and $\lambda$ that controls the permutation models need to be learned. We have assumed a Gaussian prior on $\eta$. The role of this prior is to set $\eta$ to zero when we do not observe any *significant* difference between $\pi$ and $\sigma$ i.e $\pi_i \approx \sigma_i$.

We first assume that $\lambda$ and $\mu$ are predefined constants and describe the inference (LTP-INF) of the personalization vector $\eta$ based on these values in Section 4.2. We will then use LTP-INF to also estimate these parameters in Section 4.3.

### 4.2 Inference of Personalization Vector

The key inferential problem that we study in this work is to obtain the posterior distribution on the latent variables i.e. to determine $p(\theta, K, \beta, z, \tau, \eta \mid \sigma; \lambda, \mu)$. As with simpler topic models, the exact inference is intractable and therefore, we resort to approximate inference techniques. Given

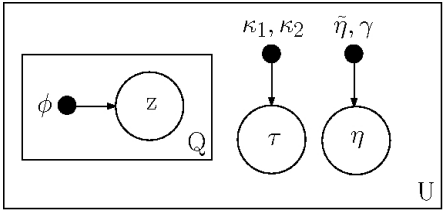

Personalization Block

**Figure 3: Variational distribution used for inferring personalization in LTP.**

the non-conjugacy of $\pi$ and $\theta$, sampling based techniques are unlikely to be efficient. In this paper, we propose a variational approximation scheme. In a variational inference, one defines a family of simpler distribution over the latent variables to approximate the true posterior distribution. This family of distribution is indexed by additional parameters (called *variational parameters*) which are tuned so as to minimize the KL divergence with the true posterior.

We first simplify the inference by breaking it into two parts. For the first part, we ignore the dependency between the topic and the personalization block. Therefore, our strategy is to first infer the topics and use the inferred topics and the topic-maps of the items to carry out inference for the personalization block. This will simplify the exposition greatly and the ideas that we develop here will carry over naturally to the general case of inferring the blocks jointly. We revisit the inference for the complete model in Section 4.4. Inference for the topic block follows standard techniques (see e.g. [3]) and therefore, we omit the details here. For the rest of this sub-section we assume that the topics have been inferred and develop an inference scheme for the personalization block.

For the personalization block, the key inferential problem is to obtain the posterior distribution $p(z, \tau, \eta \mid \sigma; \lambda, \mu)$. This posterior is approximated with the help of a variational distribution $r$. Figure 3 illustrates its graphical model representation. The personalization vector $\eta$ is assumed to be Gaussian with the following density

$$r(\eta \mid \tilde{\eta}) = (2\pi\gamma^2)^{-\frac{T}{2}} \exp\left(-\frac{1}{2\gamma^2}(\eta - \tilde{\eta})' \cdot (\eta - \tilde{\eta})\right)$$

Here, the variational parameter $\tilde{\eta}$ represents the mean of the gaussian and its variance is $\gamma^2 \mathbb{I}$. For query $q_i$ we assume that $z_i$ is sampled from a Bernoulli distribution with parameter $\phi_i \in (0, 1)$. Finally, for user $u$, we assume that $\tau$ is sampled from a beta distribution having the following density function

$$r(\tau \mid \kappa_1, \kappa_2) = \frac{\Gamma(\kappa_1 + \kappa_2)}{\Gamma(\kappa_1)\Gamma(\kappa_2)} \tau^{\kappa_1 - 1}(1 - \tau)^{\kappa_2 - 1}$$

where the parameters $\kappa_1, \kappa_2 > 0$ and $\Gamma(x)$ is the Gamma function. We use the notation $\Psi(x)$ for the digamma function which is defined as $\frac{d}{dx} \ln \Gamma(x)$.

The next step in our variational analysis is to learn the particular value of the parameters $(\phi, \kappa_1, \kappa_2, \tilde{\eta})$ that minimizes the KL divergence between $r$ and the true posterior $p$. It can be shown[10] that minimizing the KL divergence has the same effect as maximizing the following objective

---
[10]Refer to [3] for the proof.

**Algorithm 1** *LTP-INF: Variational Inference Algorithm for LTP*
---
1: **Input** Training data-set $(\pi, \sigma)_{1,2,\cdots,m}$; values for $\lambda, \gamma, \delta, \mu$;
2: **Output** Values $(\phi'_{1\ldots m}, \kappa'_1, \kappa'_2, \tilde{\eta}')$ that maximize $\Lambda$;
3: **Initialization** Randomly initialize to $(\phi^{(0)}_{1\ldots m}, \kappa^{(0)}_1, \kappa^{(0)}_2, \tilde{\eta}^{(0)})$ such that $1 > \phi^{(0)}_{1\ldots m} > 0$ and $\kappa^{(0)}_1, \kappa^{(0)}_2 > 0$;
4: $i \leftarrow 0;\quad \Delta^{(0)} \leftarrow \Lambda(\phi^{(0)}_{1\ldots m}, \kappa^{(0)}_1, \kappa^{(0)}_2, \tilde{\eta}^{(0)})$;
5: **while** $\Delta$ has not converged **do**
6: $\quad i \leftarrow i + 1$;
7: $\quad \kappa^{(i)}_1 \leftarrow \delta + \sum_{j=1}^m (1 - \phi^{(i-1)}_j)$;
8: $\quad \kappa^{(i)}_2 \leftarrow \delta + \sum_{j=1}^m \phi^{(i-1)}_j$;
9: $\quad$ **for** $j = 1\ldots m$ **do**
10: $\quad\quad \mu_j \leftarrow \Psi(\kappa^{(i)}_2) - \Psi(\kappa^{(i)}_1) + \ln f(\pi_j \mid \sigma_j, \mu) - \mathbb{E}_r[\ln g(\pi_j \mid \eta; \sigma_j, \theta, \lambda)]$;
11: $\quad\quad \phi^{(i)}_j \leftarrow 1/(1 + e^{\mu_j})$; /* Update $\phi_j$ */
12: $\quad$ **end for**
13: $\quad \tilde{\eta}^{(i)} \leftarrow \underset{\tilde{\eta}}{\text{argmax}}\ \Lambda(\phi^{(i)}_{1\ldots m}, \kappa^{(i)}_1, \kappa^{(i)}_2, \tilde{\eta})$; /* Use conjugate gradient to optimize this block */
14: $\quad \Delta^{(i)} \leftarrow \Lambda(\phi^{(i)}_{1\ldots m}, \kappa^{(i)}_1, \kappa^{(i)}_2, \tilde{\eta}^{(i)})$;
15: **end while**
16: **return** $(\phi^{(i)}_{1\ldots m}, \kappa^{(i)}_1, \kappa^{(i)}_2, \tilde{\eta}^{(i)})$
---

function,

$$\Lambda(\phi, \kappa_1, \kappa_2, \tilde{\eta}) = \mathbb{E}_r[\ln p] + \mathbb{H}(r) \qquad (4)$$

where $\mathbb{H}(r)$ is the entropy and $\mathbb{E}_r$ denotes expectation w.r.t the distribution $r$.

We use block coordinate-wise ascent to maximize the expression in Equation 4. Intuitively, we perform fixed point iterations by updating one block of parameters at a time, keeping all other parameters fixed to their most recent value. The update rule for parameters $\phi_{1,2,\cdots,m}, \kappa_1, \kappa_2$ are obtained by setting the partial derivatives of $\Lambda$ to zero. Due to our choice of $r$, the update rules for $\phi, \kappa_1, \kappa_2$ are particularly simple and have closed-form expressions.

To maximize $\Lambda$ with respect to $\tilde{\eta}$, we use the conjugate gradient algorithm[11]. The objective function for $\tilde{\eta}$ can be written as

$$L(\tilde{\eta}) = -\frac{1}{2\gamma^2} \tilde{\eta}' \cdot \tilde{\eta} + \sum_i (1 - \phi_i) \cdot \mathbb{E}_r[\ln g(\pi_i \mid \sigma_i, \theta, \lambda)]$$

It can be proved that $L$ is concave (with respect to $\tilde{\eta}$) and therefore, using simple optimizers like conjugate gradient, we will be able to obtain the global maximum [4]. Algorithm 1 summarizes the inference procedure. See Section 4.2.1 for the derivations.

### 4.2.1 Derivations

We now outline the key steps in deriving the update equations. Our first goal is to obtain the expressions for the entropy and expectation terms (Equation 4).

**Entropy of** $z$ The multinomial variate $z$ has a simple entropy expression given by $-\sum_{i=1}^m (\phi_i \ln \phi_i + (1 - \phi_i) \ln(1 - \phi_i))$.

**Entropy of** $\tau$ The entropy expression for the beta variate

---
[11]http://en.wikipedia.org/wiki/Nonlinear_conjugate_gradient_method

$\tau$ is well-known [12] and given by the following expression,

$$\ln \Gamma(\kappa_1) + \ln \Gamma(\kappa_2) - \ln \Gamma(\kappa_1 + \kappa_2)$$
$$-(\kappa_1 - 1)\Psi(\kappa_1) - (\kappa_2 - 1)\Psi(\kappa_2) + (\kappa_1 + \kappa_2 - 2)\Psi(\kappa_1 + \kappa_2)$$

**Entropy of** $\eta$ The entropy of a gaussian is a function of the covariance matrix only, which is $\gamma^2 \mathbb{I}$ and therefore a constant.

**Deriving** $\mathbb{E}_r[\ln p(z_i \mid \tau)]$ This expression requires us to determine $\mathbb{E}_r[\ln \tau]$ which can be obtained using the technique outlined in Blei et.al. [3]. Finally, the expression is given by

$$\phi_i(\Psi(\kappa_1) - \Psi(\kappa_1 + \kappa_2)) + (1 - \phi_i)(\Psi(\kappa_2) - \Psi(\kappa_1 + \kappa_2))$$

**Deriving** $\mathbb{E}_r[\ln p(\tau \mid \delta)]$ This derivation is similar to the last one and is given by

$$(\delta - 1)(\Psi(\kappa_1) + \Psi(\kappa_2) - \Psi(\kappa_1 + \kappa_2))$$

**Deriving** $\mathbb{E}_r[\ln p(\eta \mid \gamma)]$ This can be derived using standard gaussian identities [13]. The expression is simply given by $-\frac{1}{2\gamma^2} \tilde{\eta}' \cdot \tilde{\eta}$.

**Deriving** $\mathbb{E}_r[\ln(\pi_i \mid \sigma_i, \eta)]$ This derivation is more subtle. First observe that the expression is of the form $\ln(e^{\theta_1' \eta} + e^{\theta_2' \eta} + \cdots)$. This expression is in general unwieldy as the exponential terms appear inside the logarithm. We use a standard trick of simplifying this form in the following way,

$$\ln(e^{\theta_1' \eta} + e^{\theta_2' \eta} + \cdots) \leq \frac{1}{\zeta}(e^{\theta_1' \eta} + e^{\theta_2' \eta} + \cdots) + \ln \zeta - 1$$

where $\zeta$ is an additional variational parameter. Observe that the inequality holds for every $\zeta > 0$ and equality is attained only for

$$\zeta = e^{\theta_1' \eta} + e^{\theta_2' \eta} + \cdots$$

We now have to deal with the expectation term $\mathbb{E}_r[e^{\theta' \eta}]$. Observe that in the variational model, we have assumed $\eta_i$ to be independent (conditioned on $\gamma$) and therefore, this expression is equivalent to

$$\mathbb{E}_r[e^{\theta' \eta}] = \mathbb{E}_r[\prod_k e^{\theta_k \cdot \eta_k}]$$
$$= \prod_k \mathbb{E}_r[e^{\theta_k \cdot \eta_k}]$$

The expectation term can be derived using the Moment Generating Function of gaussian distribution and evaluates to $\prod_k \exp(\theta_k \tilde{\eta}_k + \frac{1}{2} \theta_k^2 \cdot \tilde{\eta}_k^2)$.

The expression for ELBO can be obtained by summing up all the entropy and expectation terms. Finally the update equations are derived by setting the partial derivatives of ELBO to zero for each block.

### 4.3 Parameter Estimation

We now focus our attention at learning $\lambda$ and $\mu$. We use Maximum Likelihood Estimators (MLE) for this, where one finds the value of the parameters that maximizes the (log) likelihood of the observed data i.e. the following expression

$$\ln p(\pi \mid \lambda, \mu; \sigma) = \sum_{i=1}^{m} \ln p(\pi_i \mid \lambda, \mu; \sigma_i) \quad (5)$$

---
[12] see http://en.wikipedia.org/wiki/Beta_distribution
[13] See www.cs.nyu.edu/ roweis/notes/gaussid.pdf

**Algorithm 2** *LTP-EM: Variational EM Algorithm for LTP*

1: **Input** Training data-set $(\pi, \sigma)_{1,2,\cdots,m}$
2: **Output** Values $(\lambda', \mu')$ that maximize Equation 5
3: **Initialization** Randomly initialize $(\lambda^{(0)}, \mu^{(0)})$ s.t. $0 \leq \lambda^{(0)} \leq 1$ and $\mu^{(0)} > 0$.
4: **while** $(\lambda, \mu)$ have not converged **do**
5:    **E-step**    /* The variational inference step */

    • $(\phi'_{1\cdots m}, \kappa'_1, \kappa'_2, \tilde{\eta}') \leftarrow$ LTP-INF$(\sigma, \pi, \lambda^{(i)}, \mu^{(i)})$;

    • $\Lambda^{(i)}(\lambda, \mu) \leftarrow \mathbb{E}_{r(\phi', \kappa'_1, \kappa'_2, \tilde{\eta}')}[\ln p]$;

6:    **M-step** /* Learn new estimates of the parameters */

    • $(\lambda^{(i+1)}, \mu^{(i+1)}) \leftarrow \underset{\substack{\mu > 0 \\ 1 \geq \lambda \geq 0}}{\operatorname{argmax}} \Lambda^{(i)}(\lambda, \mu)$

7:    $i \leftarrow i + 1$
8: **end while**
9: **return** $(\lambda^{(i)}, \mu^{(i)})$

---

However, to calculate the likelihood function, we have to marginalize over the latent variables which is difficult in our model for both real variables ($\eta$, $\tau$), as it leads to integrals that are analytically intractable, and discrete variables ($z_{1\cdots m}$), it involves computationally expensive sum over exponential (i.e. $2^m$) number of terms.

We use the variational Expectation Maximization (EM) algorithm to circumvent this difficulty. In the E-step, Algorithm 1 approximates the true posterior distribution over the latent variables, using the current estimates of the parameters. The variational parameters learned in this step are used in the subsequent M-step to maximize the likelihood function (over the true parameters $\lambda$ and $\mu$).

Algorithm 2 summarizes the steps of the variational EM. It can be shown (see Section 4.2.1) that the constraint maximization problem in step 6 is a concave program and therefore, can be solved optimally and efficiently [4].

### 4.4 Learning Topic Distributions

For inference in the topic block (Figure 1), we augment our variational distribution with additional parameters in the following way. Topic distribution $\beta_k$ is sampled from a Dirichlet prior with parameters $\{\tilde{\beta}_{k,w} \mid w \in V\}$. The topic assignments $K_{i,j}$ are sampled from a multinomial distribution with parameters $\omega_{i,j,1\cdots T}$ and $\theta_i$ is sampled from a normal distribution with mean $\tilde{\theta}_i$ and variance $\alpha^2 \mathbb{I}$. Using the same recipe as in Section 4.2 (c.f. Equation 4), we arrive at the following simple update rule for learning the topic distributions

$$\beta_{k,w} = \nu + \sum_{\substack{i,j \\ W_{i,j} = w}} \omega_{i,j,k}$$

The topic assignments $\omega_{i,j}$ also has a closed form update rule as given by $\omega_{i,j,k} \propto \exp(\mathbb{E}_r[\ln \theta_i] + \mathbb{E}_r[\ln \beta_{k,w_{i,j}}])$

Learning of topic-maps of the urls (i.e. $\theta_i$'s) is more subtle. The main difficulty stems from the coupling between the personalization and the topic blocks through $\theta$. While determining $\mathbb{E}_r[\ln g(\pi \mid \eta, \theta; \sigma, \lambda)]$ (step 8 of Algorithm 1), we now have to take expectation over $\theta$, in addition to $\eta$. Specifically, we have to compute an expectation of the form

$\mathbb{E}_r[\exp(\lambda \eta' \cdot \theta + \frac{\lambda^2 \gamma^2}{2} \theta' \cdot \theta)]$ which is however tractable due to our assumption of independence and gaussian priors on $\theta$ and $\eta$. We use gradient descent on $\theta$ to solve it. The rest of the calculation remains unchanged.

## 5. EXPERIMENTS

In this section, we describe a comprehensive set of experiments designed to evaluate the accuracy and effectiveness of our techniques.

### 5.1 Datasets

The input to our algorithm consists of a set of queries and the personalized and vanilla results (i.e. $\pi, \sigma$ pairs) for them, returned by a search engine. During the training phase, we present these queries to LTP and let it learn the personalization vector $\eta$. Once $\eta$ is learned, the next step is to validate it, by measuring how well it corresponds to the ground truth. However, in practice, such validation schemes are often difficult to design as the search engines do not reveal the actual user profile[14]. We therefore perform our experiments on both real-world dataset comprised of Google search history of a few users, and a large scale synthetic dataset.

#### 5.1.1 Google Search Personalization

We collected search result and history data[15] from 10 real Google users. This data collection was done as part of a larger survey to understand the topic level personalization and privacy concerns of users, and is part of an ongoing initiative to build a privacy evaluation toolkit (see Section 1.2). Of this larger group, due to privacy concern, only 10 participants volunteered to share their search history.

For these 10 users, we fetched their entire history of search queries. The average number of (distinct) search queries was 872. We issued each query to Google both by using their login credentials and without it to retrieve the search results. We used the Mallet [18] toolkit to extract topics from the entire collection of urls [16] returned for all queries, for each user.

We found ample evidence of profile based personalization on Google. Even when the personalized and vanilla queries were performed with identical parameters, such as location and IP address (same machine), user-agent, other http-connection, etc., roughly 30% queries received personalized results. We also found that the personalization is much more subtle compared to the impression we get from search personalization literature (and our experiments with AlterEgo server)—most queries ($\approx 70\%$) were not personalized and while there were some queries with fair amount of personalization, on an average, we observed very little difference between the results[17].

#### 5.1.2 AlterEgo

We use an open source search personalization engine called AlterEgo [17] to generate the synthetic dataset. AlterEgo contains implementation of various popular profiling and personalization techniques; we used their "unique matching" technique for our experiments[18]. In our simulation, we used AlterEgo as a surrogate personalization engine i.e. we obtain the vanilla result from Google and use AlterEgo to personalize it. The benefit of this approach is that we can train AlterEgo on topics of our choice and use this information to validate the model output $\eta$. The work-flow and details of the data generation steps are presented below.

**Generating Topics** We extracted a set of 500 topics by running Mallet on approximately 420k urls obtained from the Delicious dataset[27]. We manually select 50 topics and label them into 10 categories (examples are health, cooking, science, finance, etc.); these topics serve as a ground-truth for us. The selection of these topic categories and urls (used in the next step) is intended to simulate a typical user behavior, where, a user in interested in $\approx 10$ categories of topics.

**Training AlterEgo** For each topic, we inspect the topic-maps of the urls and identify the ones which have significant ($> 0.2$) weight (on this topic). These urls are used to train AlterEgo profile. We generated 10 profiles trained on a subset of 1 to 10 topics (i.e. 10 profile for 1 topic, 10 profile on 2 randomly selected topics, and so on), generating a total of 50 profiles.

**Queries** We generated 500 queries for each topic by randomly combining the top 10 relevant words from them. This gives us a total of 5k queries (over 10 categories). For each query, we retrieved the vanilla results from Google. Note that, if a query is related to a topic used for training the profile, only then AlterEgo will be able to personalize it. Otherwise, the vanilla and personalized results will be more or less identical.

### 5.2 Implementation Details

We implemented Algorithms 1 and 2 in the Java programming language. For solving the convex program in Algorithm 2 (step 6), we use JOptimizer [9] - a java based open source optimization package. All our experiments are carried out on a Intel Pentium IV machine with 3.0GHz processor and 4GB of RAM.

We use the following values of the hyperparameters : $\delta = 2.0, \gamma = 1.0$. For computational efficiency, we used Mallet for inference in the topic-block (see Figure 1) and do not use the inference process described in Section 4.4.

### 5.3 Results with the AlterEgo data-set

In this section, we summarize the result of our experiments with the AlterEgo data-set.

#### 5.3.1 Precision-Recall

Our first set of experiments are designed to evaluate the accuracy of LTP in correctly learning the personalized topics. On each AlterEgo profile, we train LTP and learn the personalization vector $\eta$. Next we compare it with the actual list of topics that were used to train this profile (by AlterEgo). Let $T_{act}$ be the true set of personalized topics and $T_{inf}$ be the one inferred by LTP. For this experiment, we measure the precision and recall values, where precision is defined as $\frac{|T_{act} \cap T_{inf}|}{|T_{inf}|}$ i.e. the fraction of reported topics that are actually personalized and recall by $\frac{|T_{act} \cap T_{inf}|}{|T_{act}|}$ i.e. the fraction of the original personalized topics that we are able to identify.

---

[14] Google, however, publishes the categories of topics used to serve personalized ads. Unfortunately, this data is not quite helpful as the categories are very high level and do not convey rich enough information.

[15] http://history.google.com/history

[16] We used the snippets that Google returns along with the search results to obtain text for the urls.

[17] The avg. EMD (earth mover's distance) over queries with personalization was 5.9 (e.g. the EMD of moving a single url at rank 5 to rank 1 is 4)

[18] We also did experiments with their "matching" technique, and got very similar results which are omitted due to lack of space.

| P@1 | P@3 | P@5 | R-pre | P@+1 | P@+3 | MAP |
|---|---|---|---|---|---|---|
| 97.80 | 84.02 | 70.60 | 84.66 | 70.69 | 54.44 | 97.60 |

Table 1: Performance (in %) of LTP in finding personalized topics.

We re-order the topics based on the (decreasing) value of $\eta$ computed by LTP. For each $k$, we declare the top-k topics (with maximum $\eta$ values) as personalized and calculate the precision and recall value for this decision. Table 1 summarizes the precision scores obtained by LTP. Specifically, we evaluate its performance in terms of Precision@1(P@1), P@3, P@5, R-precision (R-pre) and mean average precision (MAP) [5, 7]. Note that the size of actual topics was quite different for different runs (varies from 1-10). Hence, along with the top-k topics, we also study the precision at $|T_{act}+k|$ (denoted as P@+k).

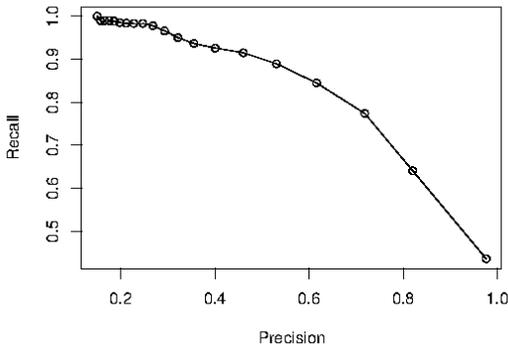

Figure 4: Precision-Recall results for LTP in retrieving the personalized topics.

In Figure 4, we illustrate the recall performance of our algorithm. At the expense of low precision ($< 0.4$), LTP is able to retrieve all the personalized topics (recall $\geq 0.93$) and its recall performance is relatively insensitive to precision; however, if we require high precision ($> 0.8$), the recall drops to $\approx 0.5$. As evident from the figure, a typical operating characteristic of LTP is precision $\approx 0.7$ and recall $\approx 0.7$, which is achieved when we return top-3 topics.

### 5.3.2 Classification Tests

In this section, we develop two classification tests to evaluate LTP's predictive power. For both these experiments, we randomly split the $\pi, \sigma$ list into data-sets D1 (80%), used for training LTP, and D2 (20%), used for testing. We repeat this split with 10 random seeds and report the average number in all the data presented below.

**Query Disambiguation** In this experiment, while testing on D2, we hide which result is personalized and which one is vanilla and the task of the model is to determine the correct labels.

We proceed with the classification task in the following way. Let $\eta'$ be the parameter learned by LTP during the training. For input lists $l_1$ and $l_2$, LTP calculates the likelihood values $p(l_1 \mid l_2, \eta')$ and $p(l_2 \mid l_1, \eta')$ and whichever likelihood is higher is assigned to the personalized result i.e. if $p(l_1 \mid l_2, \eta') > p(l_2 \mid l_1, \eta')$ then $l_1$ is declared to be the personalized result and vice versa. We name this test as *P-V disambiguation* for a given profile. Over all the test points in D2, the fraction of queries that were labeled correctly is referred to as *disambiguation accuracy*.

| #topics | Accuracy ($\mu \pm \sigma$) | | Time (secs) | |
|---|---|---|---|---|
| | LTP-EM | LTP-INF | LTP-EM | LTP-INF |
| 1 | .74 ± .09 | .72 ± .09 | 80.7 | 22.7 |
| 2 | .72 ± .06 | .70 ± .09 | 154.3 | 31.5 |
| 3 | .70 ± .05 | .68 ± .06 | 221.6 | 42.4 |
| 4 | .69 ± .04 | .67 ± .05 | 272.2 | 53.7 |
| 5 | .69 ± .05 | .67 ± .05 | 336.1 | 69.8 |
| 6 | .67 ± .04 | .65 ± .05 | 333.2 | 70.7 |
| 7 | .65 ± .04 | .65 ± .05 | 342.5 | 71.1 |
| 8 | .63 ± .04 | .63 ± .04 | 348.2 | 73.6 |
| 9 | .63 ± .05 | .62 ± .05 | 354.4 | 76.4 |
| 10 | .62 ± .02 | .62 ± .02 | 359.2 | 79.5 |

Table 2: Summary of results with the AlterEgo dataset

Table 2 summarizes the result of this experiment. In summary, we achieve disambiguation accuracy in the range of 62-74%. For each profile, we collect the accuracy values for the 10 different runs and report its mean and standard deviation ($\mu \pm \sigma$). Observe that our accuracy decreases slightly as the AlterEgo profile is trained with more and more topics.

Table 2 also reports the training time of LTP-EM. For profiles trained with many topics, LTP-EM takes more time to converge. We repeat the experiment with LTP-INF with the parameter values fixed to $\lambda = 0.9$ and $\mu = 10.0$. As the results show, LTP-INF is up to 5 times faster to train but achieves slightly lower accuracy. The accuracy however, improves slightly ($< 3\%$) if we increase the amount of training data (D1) from 80% to 90% (not shown in the table).

**User Classification** For this experiment, we consider groups of users (i.e. profiles) and develop a classification test within the group members. We vary the size of the group from 2 to 10 and for each group size, randomly pick 10 groups. For each group $G$, we present a $(\pi, \sigma)$ pair to LTP but do not reveal the user it belongs to. The task of the model is to correctly predict the user.

We again use the likelihood test for this task. Specifically, for each user $u \in G$ and input $(\pi, \sigma)$, we calculate $p(\pi \mid \sigma, \eta'_u)$ ($\eta'_u$ learned during training) and output the user for which the likelihood attains its maximum value.

In Figure 5, we summarize the result of this experiment. There are two parameters in this experiment - the size of the group and the number of topics used to train AlterEgo for each profile in the group. For simplicity, we present here results for the homogenous case, where we combine profile which are trained on the same number of topics [19]. Observe that the accuracy reported by LTP is significantly higher than a random guess (which is 1/g, g being the group size). The accuracy decreases slightly if profiles are trained with many topics. We believe this reduction in accuracy is also an artifact of our data generation—profiles trained on multiple topics can (and do) have topics in common, that will make it hard to distinguish personalized response on two profile trained on the same topic.

---

[19]We also performed experiments on the general case (e.g. by grouping profiles trained on 3 topics with 5 topics). The results are similar and not repeated here.

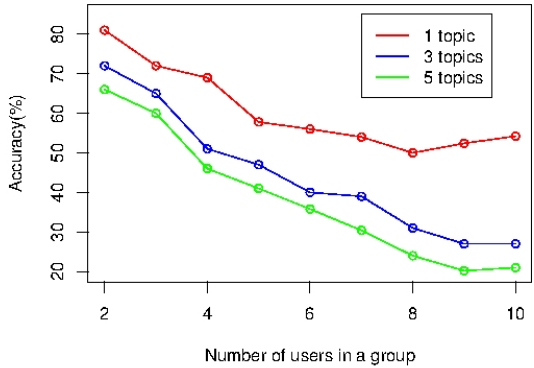

**Figure 5: Performance of LTP in user classification.**

In summary, these results, together with the precision-recall values from last section highlight that our model fits the data well and learns the correct set of personalized topics on synthetic data.

## 5.4 Results with the Google dataset

In this section we describe the results with the Google dataset. Note that since we do not know the actual personalization on different topics (ground truth) for a real Google user, we cannot perform the precision-recall experiments as with AlterEgo dataset, and resort to only query disambiguation and user classification test described above. However, we also perform some qualitative tests that give ample indication that we have found a good personalization vector.

### 5.4.1 Qualitative Evidences for Correctness of $\eta$

We now present our analysis on finding qualitative correctness of $\eta$ using *evidences* of personalization. An evidence is an instance of $\pi, \sigma$ where results were re-ranked such that the ones with $\eta$ were moved up. Note that while such evidence have no statistical significance, they are much more helpful for a user's understanding of his profile compared to the personalization vector. Such evidences are a core feature of the privacy toolkit we are building (see Section 1.2).

Figure 6 shows an example evidence of personalization happening on a user's account. The result for query Q ("how to decide mixing of markov chain") and theta values for two relevant topics T1 (about "Algorithms" defined by words algorithm, design, complexity) and T2 (about "Probability" defined by words probability, distribution) are shown. For this user, $\eta$ value for T1 is very high compared to T2. Observe that the wiki link U1 (in the box), although less relevant to the query, is placed higher in the personalized results. As our analysis shows, U2 is has a high weight on topic T1 compared to U2, which leads to this personalization. The user can therefore see not just his inferred interests (more in "Algorithms" compared to "Probability"), but also *how* it affects his results.

We next move to another qualitative analysis of $\eta$ by comparing it directly with the categories Google itself associates with a user[20]. We try to match topics with high $\eta$ (top-k such topics) with the broad categories in Google. Table 3 shows the result of such matching for 3 users. Take for example, the "Anime and Manga" category, that was also assigned a very high $\eta = .6$ (compared to an average value of .004) by LTP.

Such anecdotes show that our techniques have, in fact, learned the personalization vector correctly.

### 5.4.2 Quantitative Experiments

**Query Disambiguation** Table 4 summarizes the result of query disambiguation on the Google dataset. We first study the effects of number of topics ($T$) chosen for the user. We notice that only a few topics 15-50 are enough to get good accuracy for any user. Our accuracy results differ significantly for different users, varying from as low as 54% to 85%. We believe this is because the amount of personalization is different for various users, and this affects the learning accuracy of our techniques.

**User Classification** Table 5 show that even with 3 users, we are able to get an accuracy of up to 60%. For this experiments, we extracted $\eta$ values over a common set of topics for each user. These $\eta$ values learned were also very different for different users (data not shown). This shows that $\eta$ is in fact learned tailored to the personalization of each user.

## 6. CONCLUSIONS

In this paper we have presented a novel approach to extract user profile information in the form of personalization vector over topics from commercial OSPs (such as Google search). Our approach treats OSPs as black-boxes, i.e. assumes no knowledge of the personalization algorithms and history of users maintained by them, and works by comparing the personalized and vanilla content served by them.

| Google Category | Topic in LTP | $\eta$ |
|---|---|---|
| Comics & Animation - Anime & Manga | online read manga kyojin shingeki chapter | 0.60 |
| Autos & Vehicles - Vehicle Shopping | car india chrysler price jaguar sport bmw | 0.42 |
| Computers - Software Utilities | class import common org public implement | 0.15 |
| World Localities - South Asia | seoul citi hotel location shop mall coex | 0.13 |

**Table 3: Correlation between personalized topics in LTP and Google categories.**

| User Id | 15 Topics | 20 Topics | 50 Topics | 100 Topics |
|---|---|---|---|---|
| 1 | 74±5 | 70±5 | 70±6 | 73±4 |
| 2 | 68±5 | 70±4 | 70±4 | 65±3 |
| 3 | 67±13 | 72±14 | 67±13 | 73±11 |
| 4 | 54±8 | 51±6 | 55±6 | 59±7 |
| 5 | 54±11 | 47±9 | 49±11 | 43±9 |
| 6 | 85±7 | 78±5 | 84±4 | 81±7 |
| 7 | 73±4 | 70±5 | 71±6 | 73±6 |
| 8 | 66±3 | 62±3 | 61±4 | 64±3 |
| 9 | 52±4 | 52±3 | 50±4 | 54±4 |

**Table 4: Accuracy of LTP over 9 Google users.**

---
[20]Shown in Google ads preference manager https://www.google.com/settings/ads/onweb/

| Group Size | Number of Topics | | | |
|---|---|---|---|---|
| | 10 | 20 | 50 | 100 |
| 2 | .59 ± .06 | .61 ± .06 | .65 ± .04 | .58 ± .05 |
| 3 | .48 ± .04 | .53 ± .05 | .60 ± .05 | .50 ± .06 |

**Table 5: User classification accuracy on Google data.**

Figure 6: An example to illustrate the difference between personalized (left) and vanilla (right) search results (for a real user) returned by Google.

|  | Topic | |
|---|---|---|
|  | T1 | T2 |
| $\eta$ | .90 | .40 |
| Q | .01 | .30 |
| U1 | .15 | .10 |
| U2 | .01 | .25 |

To the best of our knowledge, this is the first work that tries to extract information based solely on mining the output of OSPs. This aspect of our work make it unique and is beneficial in not just enabling access to (so far hidden) profiles in OSPs, but also in providing a novel and practical approach for retrieving information from OSPs by mining differences in their outputs.

Our approach also has direct benefits for end users, as it for the first time, enables them to access their (so far hidden) profile information tracked by an OSP. While being an informational tool by itself, this has wider implications to the outlook of user privacy research—it can be used to infer the personalization happening on *sensitive* topics (e.g. financial, medical history, etc.), which a user may not be comfortable with. We believe that this can be used to build an end-user privacy perserving tool and are currently working on a prototype for the same.